\documentclass[runningheads]{llncs}
\usepackage[T1]{fontenc}
\usepackage{amsmath}
\usepackage{amssymb}
\usepackage{graphicx}
\vspace{-10cm}
\begin{document}
%
\title{ExtSwap: Leveraging Extended Latent Mapper for Generating High Quality Face Swapping}
%
%
\author{Aravinda Reddy PN\inst{1}\orcidID{0000-0002-1342-924X} \and
K.Sreenivasa Rao\inst{1}\orcidID{0000-0001-6112-6887} \and
Raghavendra Ramachandra\inst{2}\orcidID{0000-0003-0484-3956}\and
Pabitra Mitra\inst{1}\orcidID{0000-0002-1908-9813}
}
\authorrunning{Aravinda et al.}
%
\institute{Indian Institute of Technology Kharagpur, India \\ \email{aravindareddy.27@iitkgp.ac.in} \and
Norwegian University of Science and Technology (NTNU), Gjøvik, 2815, Norway. \email{raghavendra.ramachandra@ntnu.no }
}
\maketitle              

\begin{abstract}
We present a novel face swapping method using the progressively growing structure of a pre-trained StyleGAN. Previous methods use different encoder–decoder structures, embedding integration networks to produce high-quality results, but their quality suffers from entangled representation. We disentangle semantics by deriving identity and attribute features separately. By learning to map the concatenated features into the extended latent space, we leverage the state-of-the-art quality and its rich semantic extended latent space. Extensive experiments suggest that the proposed method successfully disentangles identity and attribute features and outperforms many state-of-the-art face swapping methods, both qualitatively and quantitatively.

\end{abstract}
\section{Introduction}
Nowadays there is obsolete rise in manipulated images, videos and audio because of the increasing elegance of camera technology and the accelerating reach of social media platforms have enabled the creation of images and videos more sophisticatedly. Before the advent of machine learning and computer vision techniques, many fake videos and their degree of resemblance to the original needed to be improved owing to a lack of domain expertise and sophisticated manual editing tools. However, with the dawn of machine learning, manual editing of videos and images has been drastically reduced.

The evolution of an Artificial Intelligence-based fake video generation method, called  Deepfakes \cite{korshunov2018deepfakes}, has lured many masses. It takes input of a video/image of a source subject and a video/image of a target subject and outputs a swapped video/image consisting of identity features of the source subject and attributes of the target subject. The backbone for the creation of these deep fakes is Deep Neural Networks (DNNs). These DNNs are trained on facial images to map facial expressions from the source to the target images. Most of the aforementioned studies use generative adversarial networks (GANs), Autoencoders (AE), and generative normalization flows (Glows) to generate images and videos as deepfakes. Among the three networks, GANs are very popular for generating indistinguishable deepfakes.

In past years, generative adversarial networks (GANs) have been used for high-quality image synthesis, particularly for facial manifolds. In particular, StyleGAN \cite{karras2019style,karras2020analyzing} proposed a style-based generator architecture and synthesis of state-of-the-art face images. One of the core problem of machine learning is to learn the disentangled representation. Disentanglement is breaking down or disentangling each feature into narrowly defined variables and encoding them as separate dimensions. GANs assemble the latent codes of the source and target image to generate the swapped image. However, the latent codes are highly entangled, and directly assembling them may not guarantee the transfer of source and attribute features onto swapped face. We argue that identity and attribute features should be separately encoded and transferred onto the swapped face. So a proper disentanglement of identity and attributes are required.

With the expeditious evolution of GANs, numerous methods have demonstrated the capability of controlling the latent space of $w$ \cite{shen2020interpreting}, which has a disentanglement property that offers control and editing capabilities. These above mentioned works follow \textit{"invert first, edit later"} approach(input image is invertible, i.e., there exists a latent code that can reconstruct the image, and with latent manipulations, the image can be edited), wherein the image is inverted into StyleGAN's latent space and then the latent code is edited in a conceptually meaningful form to obtain a new latent code and then this edited latent code is used by StyleGAN to generate the high-resolution image. However, inverting a real image into a 512-dimensional vector $w \in\mathbb{W}$ does not synthesize the image accurately. Inspired by this, many researchers \cite{abdal2020image2stylegan++} proposed an extended latent space $ w \in \mathbb{W^+}$. The extended latent space is a concatenation of 18 different 512-dimensional \textit{w} vectors, and each vector is fed as input to Adaptive Instance Normalisation (AdaIn) of StyleGAN. These studies map an image onto the extended latent space, which requires several minutes for an image to synthesize. To expedite this optimization process, they \cite{abdal2020image2stylegan++} proposed a domain-guided encoder and used it for further domain-regularized optimization. However, fast and precise inversion of real images in $\mathbb{W^+}$ remains challenging. 


This paper proposes a new framework for face swapping called \textit{ExtSwap}. We extend the work of \cite{nitzan2020face} by passing the encoded features to the extended latent space of StyleGAN; we show that our method outperforms the current state-of-the-art methods both qualitatively and quantitatively. To accomplish this task, we use two encoders that encode the identity and attribute embeddings separately. The identity encoder $E_{id}$ is the pre-trained face transformer \cite{zhong2021face} where modified tokens, and image patches, were made to overlap and projected onto the transformer encoder. For the attribute encoding, we use EfficientNetV2 \cite{tan2021efficientnetv2}. EfficientNetV2 is well-known for faster training and improved parameter efficiency than EfficientV1 \cite{tan2019efficientnet}.  In the former, am automatic way of deriving architecture using Neural Architecture Search (NAS) and compound scaling of depth, width, and image size was performed with a fixed ratio. The features generated from these two encoders are concatenated and fed into four layer Multi-Layer Perceptron (MLP). The characteristic of MLP is that it can learn from nonlinear models. Because two different encoders encode the image and provide two other dimensional tensors to map them and feed into StyleGAN's generator, an MLP is used. The mapped style vectors generated from the MLP were inserted into a pre-trained StyleGAN generator. To generate a swapped face, the style vectors are encoded into extended space $\mathbb{W^+}$. 
The overall contribution of this paper is as follows:

\begin{enumerate}
    \item A novel framework for high-resolution face swapping called ExtSwap. A novel method for disentangling identity and attribute features.
    \item A new encoders for encoding and concatenating the identity and attribute features of facial images.
    \item We compare our method with several other state-of-the-results and show that our method outperforms the current state-of-the-art methods.
\end{enumerate}

\section{Related work:}
\subsection{Face Swapping:}
The first ever face swapping method was introduced by \cite{bitouk2008face} by aligning both target and source images to a common coordinate system and then by replacing the six fiducial points of the source with the target. However, this method required the posture of the target face to be identical. Later \cite{korshunova2017fast} used CNNs to transfer a person's identity while keeping head, pose, identity, and lighting conditions intact. To pragmatically puppeteer and animate a face from a single RGB image, a 3D model is used to extract the textures and to transfer the textures onto target image GANs were used \cite{olszewski2017realistic}. A parametric and data-driven model was proposed by \cite{sun2018hybrid} wherein a 3D mophing model (3DMM) model was used in the first stage, and in the later stage, GANs were used to synthesize the full image. In another work \cite{bao2018towards}, two different encoders were used to extricate the identity and attribute of the face image, to generate a new image using GAN. A novel integrated editing system for face and hair was proposed by \cite{natsume2018rsgan} wherein they used a separator network to separate the hair and face and a composer network that reconstructs the input face from two latent spaces and a discriminator that discriminates the real and fake sample. Face SwapGAN (FSGAN) \cite{nirkin2019fsgan} presented an RNN-based approach for face reenactment. They used a fully supervised method in three stages, reenactment, inpainting, and blending. This method yielded good results for face and hair occlusions but failed for glasses, veils, and so on. A high fidelity and occlusion aware face swapping technique was introduced by \cite{li2019faceshifter}. Recently,  \cite{naruniec2020high} proposes an encoder that swaps faces at high resolution, but their model needs to be subject-agnostic. MegaFS \cite{zhu2021one} designed an id-transfer block to combine their latent codes. However, the structure and appearance part of the latent codes are entangled in the latent space.

\subsection{Latent Space of GAN}

Many researchers in the past have attempted to understand and regulate the latent space of GAN since their origin. GAN inversion intends to invert the image back into the latent space of a pre-trained StyleGAN model. The image was faithfully reconstructed using the inverted code. Several methods, such as \cite{huang2018introvae}, thoughtfully train a variational autoencoder (VAE) to mimic that of conventional GAN, and parallel VAE is adversarially trained. Another work that mimics GAN was where they designed two latent encoders that are trained adversarially and can synthesize faces and bedroom images comparable to GAN \cite{pidhorskyi2020adversarial}. Training encoders, such as GAN, often compromise the image quality and require a long training time. As a result, other methods use inversion as an alternate method, where the generator is pre-trained, and inversion is performed separately. The methods mentioned above either optimize the latent vectors to curtail the reconstruction error or train different encoders to improve the inversion efficiency \cite{richardson2021encoding}. Recently, several studies have proposed methods for latent-space manipulation. The images generated from StyleGAN are at high resolution, but there is no semantic control over the output, such as expression, pose, and illumination. Stylerig, proposed by \cite{tewari2020stylerig}, allows control over the pose, expression, and illumination parameters. In another study, \cite{nitzan2020face} utilized a disentangled property to represent data with minimal supervision by a pre-trained StyleGAN.

\section{Proposed Method}
\subsection{Overview}
\begin{figure}[!t]
 \centering
 \includegraphics[width=\textwidth]{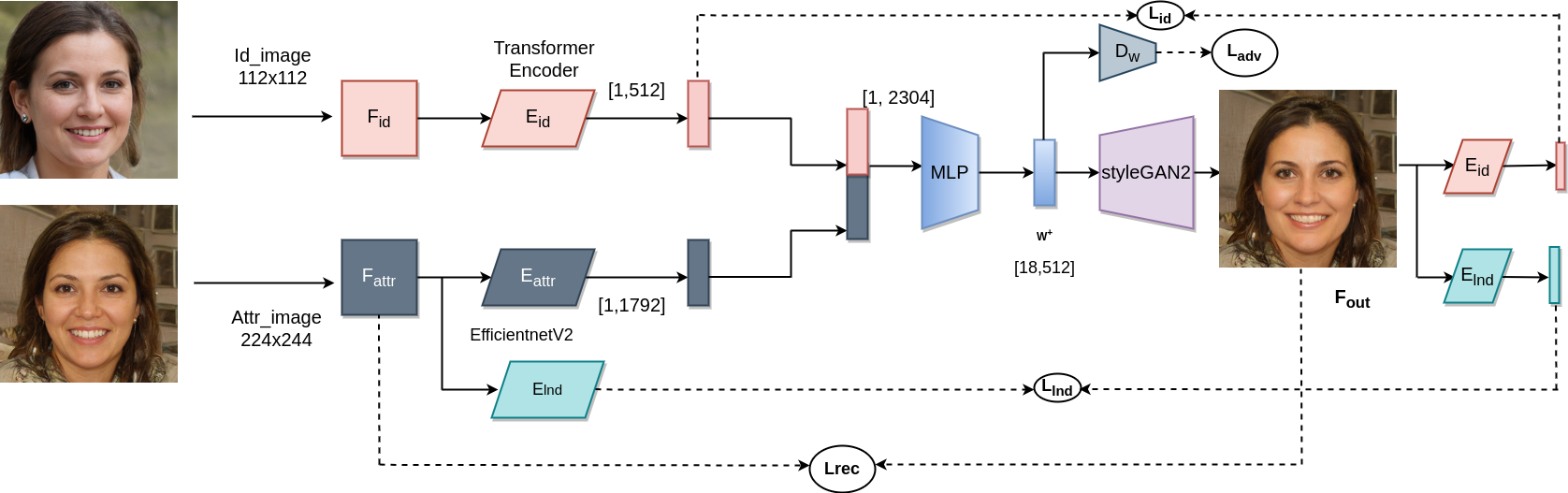}
 \caption{ExtSwap framework uses two different encoders to generate the latent code z. The code is then mapped into the pre-trained generator's extended space $\mathbb{W^+}$. Losses are marked in the dashed lines, and data flow in the solid line.}
 \label{fig:our_arch}
\end{figure}

ExtSwap takes two inputs $F_{id}$ and $F_{attr}$; we aim to construct the face image with identity features of source image $F_{id}$ and the attributes of the target image $F_{attr}$ such as pose, expression, and background. To do so, we first excerpt the identity features of source image $F_{id}$ explained in sec \ref{id_fea} and the attribute features of target image $F_{attr}$ are explained in sec \ref{attr_fea}. Identity and attribute features are concatenated and fed to the latent mapper, explained in sec \ref{lat-map}. The latents are fed to the StyleGAN to synthesize the swapped image, which is explained in sec \ref{style}. The proposed architecture pipeline is shown in Fig. \ref{fig:our_arch}

\subsection{Face Transformer} \label{id_fea}
To excerpt the identity features of the source image, we utilize the Face transformer \cite{zhong2021face} we denote it as $E_{id}$. The face transformer architecture follows the exact structure of Vision Transformer \cite{dosovitskiy2020image}, but the token generation method is modified compared to the standard ViT. The sliding patches are used to generate the tokens, where the image patches are made to overlap to preserve the inter-patch information to generate 512-dimensional embeddings. Face transformer gives comparable results to the standard ArcFace embeddings and VGG2 based Face recognition \cite{deng2019arcface}.

\subsection{EfficientNet-V2}\label{attr_fea}
To extract the attribute features of the target image $F_{attr}$, we use modified efficientnet \cite{tan2019efficientnet}. The Efficientnet is a family of models optimized for floating point operations per second (FLOPS) and parameter efficiency. It clouts the NAS to search the baseline model B0 with better accuracy and trade-off and then applies compound scaling to obtain the B1-B7 family. This method proposes an upgraded method of progressive learning which adapt the image size and regularisation. This model has an 11x faster training speed when compared to the first version. We use this encoder to extract the attributes of a facial image. A total of 1792 features were extracted from a facial image, including pose, lightning, expression, and head posture, and this encoder is referred to as $E_{attr}$.

We concatenate the codes from two encoders $E_{id}$ and $E_{attr}$ as shown in the below equation \ref{eq:zspace}
 \begin{equation}
\centering
    z_{space}=[E_{id}(F_{id}),E_{attr}(F_{attr})]
    \label{eq:zspace}
\end{equation}

\subsection{Latent Mapper}\label{lat-map}
The identity features extracted from the Face Transformer and the attribute features extracted from EfficientNet-V2 are fed to a latent mapper that acts as a dimensionality-reduction network. The 512-dimensional  identity and 1792 dimensional attribute features were concatenated to form a 2304 feature vector. This feature vector is fed to a 4-layer Multi-layer Perceptron with leaky ReLU activation layers. In this stage, 18 styles are injected in the training phase to form $(18,512)$-dimensional vector that resembles $\mathbb{W^+}$ latent space. Eighteen style vectors were extracted for a batch of images using the following equation:

 \begin{equation}
     style_vectors=log_2(resolution)*2-2
 \end{equation}
 where resolution=image resolution is $1024 \times 1024$

\subsection{Discriminator}
Since the training the mapping between the extended latent space and $\mathbb{W^+}$ is cliche. Therefore, we add Discriminator $D_\mathbb{W^+}$ to help predict the MLP predict the features lie within $\mathbb{W^+}$. So, in order to discriminate between real samples of $\mathbb{W^+}$ and MLPs predictions. we train the Discriminator in an adversarial manner.

\subsection{StyleGAN2 Generator} \label{style}
ExtSwap is based on the synthesizing power of StyleGAN2 generator G and the extended space of $\mathbb{W^+}$. StyleGAN2 improves the perceptual quality by proposing improved weight demodulation, path length regularization, by altering the generator architecture, and shedding the progressive growth compared to the first version of StyleGAN. The concatenated vector $\mathbb{W^+}$ is fed into the extended space, usually employed in the AdaIn layers to obtain the swapped face given by equation \ref{eqn:fswap}.

\begin{equation}
\centering
    F_{swap}=G(MLP([E_{attr}F_{attr},E_{id}F_{id})])
    \label{eqn:fswap}
\end{equation}

\subsection{Training and loss functions:}
We created 70,000 images by passing random Gaussian vectors into the pre-trained styleGAN2. The generated images were projected onto the extended space $\mathbb{W^+}$. We save the image and corresponding $\mathbb{W^+}$ vector. The StyleGAN generated images were used as the training dataset, and the latent $\mathbb{W^+}$ vectors were used as real samples for training the Discriminator. For training the Discriminator, we use a gradient regularization technique named R1 regularization that acts only on real data along with non-saturating loss

\begin{equation}
    \mathbb{L}_{adv}^D=\frac{1}{2}(L_{D,1}+L_{D,2})=-\frac{1}{2}[\mathop{\mathbb{E}}_{w\sim \mathbb{W^+}}[log(D_\mathbb{W^+}(\mathbb{W^+}))]]-\frac{1}{2}\mathop{\mathbb{E}}_{z \sim MLP} log[1-D_\mathbb{W^+}(MLP(z))]
\end{equation}

\begin{equation}
    \mathbb{L}_{adv}^G=\frac{1}{2}\mathop{\mathbb{E}}_{z\sim MLP}log[1-D_\mathbb{W^+}(MLP(z))]
\end{equation}

To preserve the input identity we measure cosine similarity between source $F_{id}$ and swapped image $F_{out}$, 
\begin{equation}
\mathbb{L}_{ID}=1-<R(x),R(E_{ID}(x))>
\end{equation}
 where R is the pre-trained ArcaFace \cite{deng2019arcface} Model.

To model the motion of face, we use Facial landmarks. So we use $L_2$ loss to model attribute information $F_{attr}$ to be consistent with $F_{out}$.
\begin{equation}
    \mathbb{L}_{lnd}=\|E_{lnd}(F_{attr})- E_{lnd}(F_{out})\|_2
\end{equation}

To perpetuate pixel-level information such as color and illumination, we adopt mix loss and use a weighted sum of $L_1$ loss and Multi scale Structural Similarity Index (MS-SSIM) loss:
\begin{equation}
    \mathbb{L}_{mix}=\alpha(1-MS-SSIM(F_{attr}))+(1-\alpha||F_{attr}-F_{out}||_1)
\end{equation}

However, the $F_{out}$ should not only contain the attribute information of $F_{attr}$ and also the identity information $F_{id}$ and both should be well confined. So we consider reconstruction loss only when $F_{id}=F_{attr}$, i,e., 
\begin{equation}
    \mathbb{L}_{rec}=\left\{ 
  \begin{array}{ c l }
    \mathbb{L}_{mix} & \quad \textrm{if } F_{id}=F_{attr} \\
    0                 & \quad \textrm{otherwise}
  \end{array}
\right.
\end{equation}

The overall generator loss the weighted sum of above losses:
\begin{equation}
    \mathbb{L}_{total}=\lambda_{id}\mathbb{L}_{id}+\lambda_{attr}\mathbb{L}_{attr}+\lambda_{rec}\mathbb{L}_{rec}
\end{equation}

\section{Experiments}
\subsection{Implementation Details}
We used StyleGAN pre-trained with $1024 \times 1024 $ resolution for all our experiments. We trained the attribute encoder and latent mapper only during the training, and the remaining components, such as the identity encoder and StyleGAN, were frozen during training. The ratio of training of $F_{id}$ and $F_{attr}$ is a hyper parameter that controls the disentanglement. We optimize the adversarial loss $ L_{adv}^G$ and non-adversarial loss separately $L_{non-adv}^G$. We used the Adam optimizer with $\beta_1=0.9$ and $\beta_2=0.999$ and learning rate of $2e^{-5}$ when optimizing $L_{non-adv}^G$ and $6e^{-6}$ for adversarial loss $ L_{adv}^G$ and loss weights are set to $\lambda_1=1$, $\lambda_2=1$, $\lambda_3=0.0001$ and $\lambda_4=0.02$, $\alpha=0.84$. To calculate the landmarks, we used the standard 68-point facial points. The model was implemented using a Pytorch and trained using paramshakti supercomputer which has 22 nodes, each has two GPUs of 16GB named V100 Tesla and we used one node for training. \footnote{https://github.com/Aravinda27/ExtSwap}.

\begin{figure}[!t]
\begin{minipage}[!b]{.4\textwidth}
\centering
\includegraphics[scale=0.25]{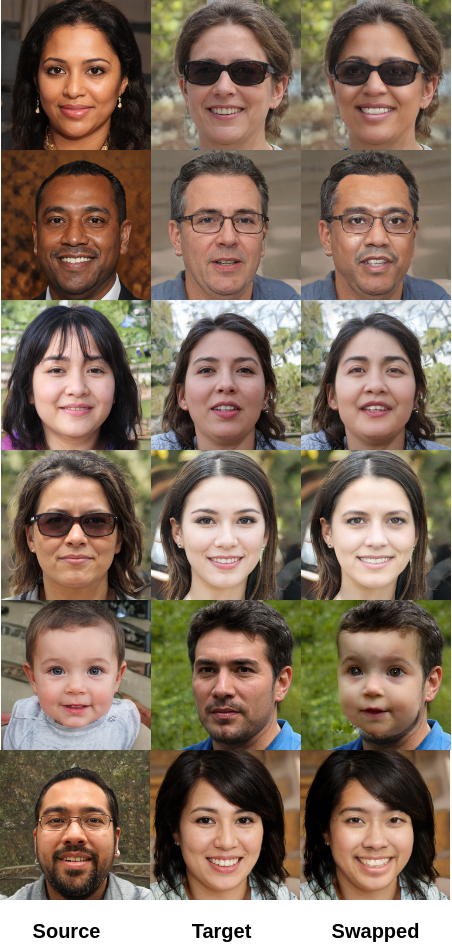}
\centering
\caption{ExtSwap results on our test data}
\label{fig:cmp_result_our}
\end{minipage}
\begin{minipage}[!b]{.6\textwidth}
\centering
\includegraphics[scale=0.25]{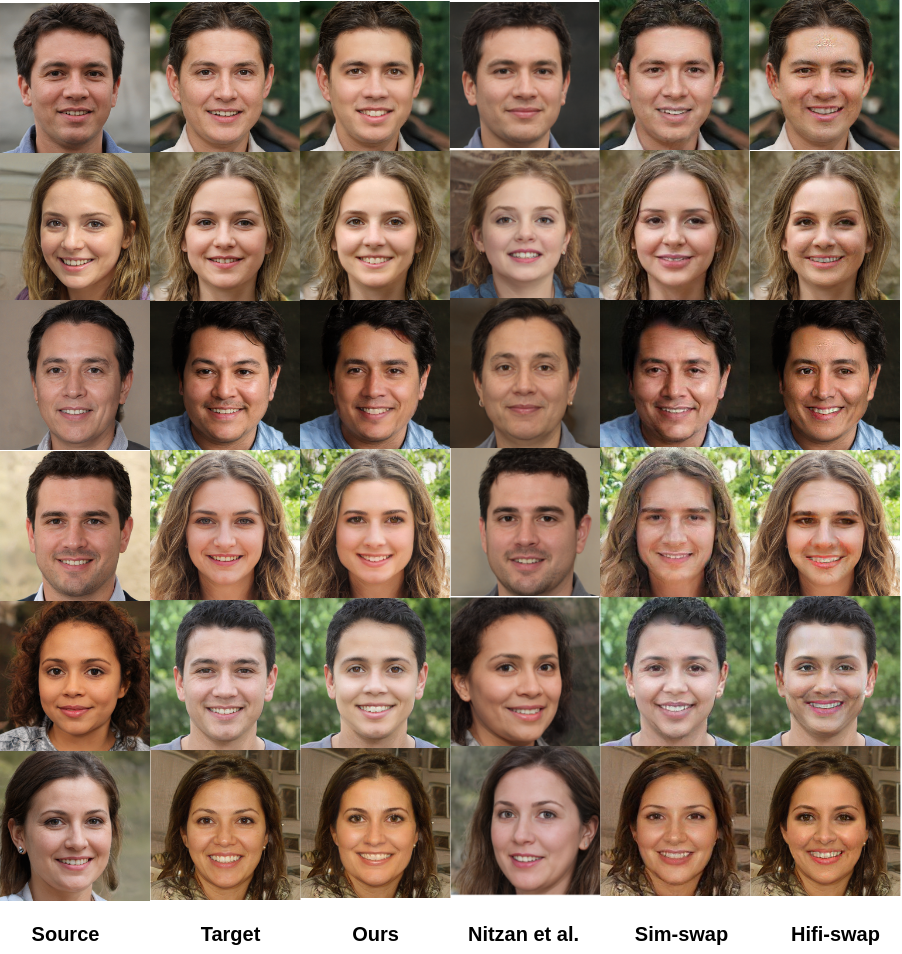}
\caption{ExtSwap compared with state-of-the-art methods}
\label{fig:cmp_results_om}
\end{minipage}

\end{figure}


\subsection{Datasets}
\textbf{Datasets}: We evalaute our model on 3 datasets:
\begin{enumerate}
\item Our test data: We split our 70,000 images into training and test sets in 80:20 manner and create random source and target pair and obtain the swapped images.
    \item CelebA-HQ \cite{liu2015faceattributes} contains 30,000 celebrities faces with high-resolution faces. We use this database for testing our model. First, we pass these images to get $\mathbb{W^+}$ vectors and create swapped faces by passing the source and target images.
    \item FFHQ dataset \cite{karras2019style} contains 70,000 high-resolution $1024 \times 1024$ face images. First, we pass these images to get $\mathbb{W^+}$ and record these vectors. We passed a batch of images by forming source and target image pair and obtained swapped images.
\end{enumerate}

\textbf{Evaluation Metrics}: To evaluate the proposed method, we calculate cosine similarity by computing the Arcface embeddings between the source and swapped image so that the swapped image contains identity features. And also we calculated the pose and expression error, which is the $l_2$ distance between the target face and swapped face, using the pre-trained model \cite{chaudhuri2019joint,ruiz2018fine}, which shows the ability to transfer the attributes. The \textit{Frechet Inception Distance} calculates the Wasserstein-2 distance between swapped faces and real faces, indicating the swapped faces' quality.

\subsection{Experiments on test data}

We conduct experiments to evaluate ExtSwap mainly on two aspects: how well the identity and attribute features appear on swapped faces and the quality of the synthesized image.

\textbf{Qualitative comparison}:
First, a qualitative inspection of ExtSwap is shown in Fig \ref{fig:cmp_result_our}. ExtSwap shows identity and attribute preservation in swapped images. In addition, it can be used to generate the overall shape of the head and hair. Moreover, we observed a consistent transfer of attributes, such as smiles and glasses, in the swapped faces.
\vspace{0.1cm}

\textbf{Quantitative comparison}: 
We performed a quantitative comparison of the proposed methods. ExtSwap has an identity similarity of 0.5698, pose error of 3.477, expression error of 2.95, and FID score of 9.87. ExtSwap shows stronger identity preservation and attribute transfer while showing lower FID scores.

\begin{figure*}[!t]
\begin{minipage}[!b]{.50\textwidth}
\centering
\includegraphics[width=0.70\textwidth]{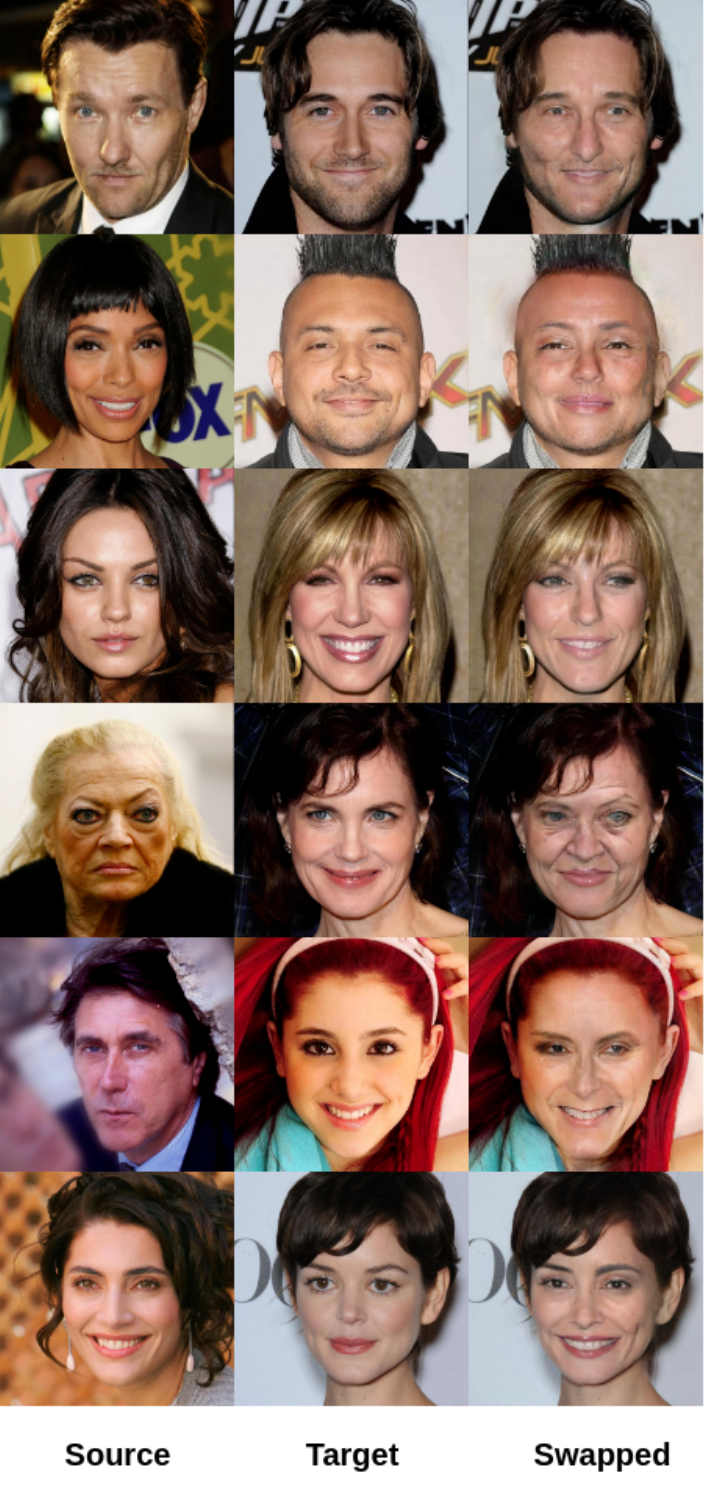}
\centering
\caption{ExtSwap results on CelebA-HQ dataset}
\label{fig:cmp_our_result_celeba}
\end{minipage}
\begin{minipage}[!b]{.50\textwidth}
\centering
\includegraphics[width=0.70\textwidth]{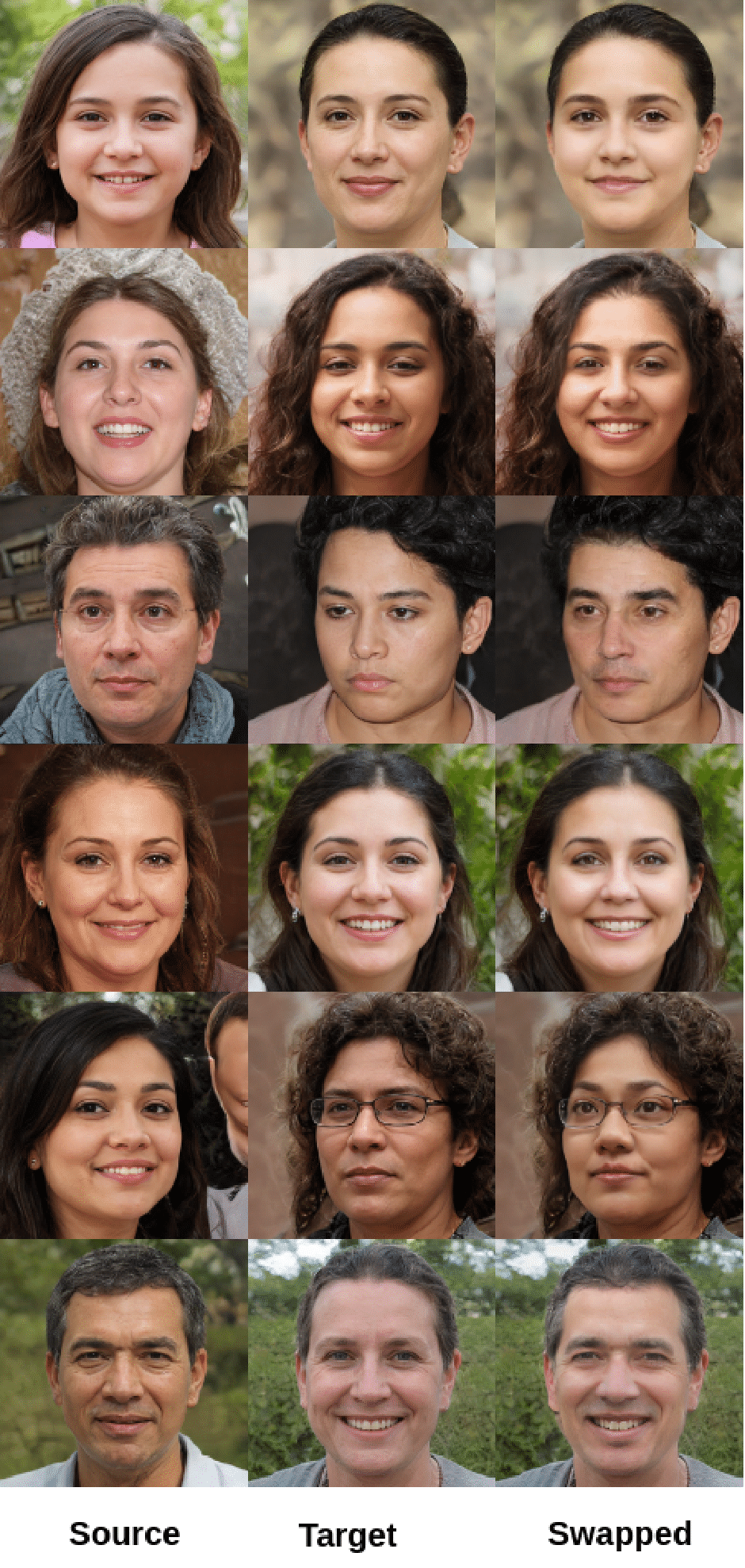}
\caption{ExtSwap compared with FFHQ dataset}
\label{fig:cmp_results_FFHQ_new}
\end{minipage}

\end{figure*}





\subsection{Comparison with previous methods}
\textbf{Qualitative comparison}:
A qualitative comparison of the proposed method is shown in Fig \ref{fig:cmp_results_om}. We compared ExtSwap with Nitzan et al.\cite{nitzan2020face}, who projected the images into a latent space $w$ rather than the extended latent space $\mathbb{W^+}$. They used VGGFace2 encoder for identity features extraction and Inception V2 network for excerpting the attribute features. VGGface uses 16 CNN layers in a sequential manner. The main disadvantage CNN networks are they do not cover long range dependencies of an image. But transformers uses self-attention thus making connection between distant image pixels. EfficientNet-V2 trains faster than inception since it uses NAS for training and thus extracting better attribute features. Also the our extended latent space has 18 styles each of 512-dimensional features which are encoded before feeding into the StyleGAN.  Since it is evident from both qualitative (Fig. \ref{fig:cmp_results_om}) and quantitative analysis ExtSwap achieves better result than \cite{nitzan2020face}.
We also compare ExtSwap with Simswap\cite{chen2020simswap} where this framework has three parts: an encoder to extract the features of the target image, the Identity Injection Module (IIM), which transfers identity information from the source to the swapped, and the decoder part, which synthesizes the image. As shown in the figure, Simswap can transfer the identity features but fails to transfer the attribute features, leading to a blurred image. Next, we compare with another technique called HifiSwap \cite{wang2021hififace}, which uses four parts: the encoder part, where features of the target image are excerpted, and a decoder that fuses the encoder and 3D shape-aware identity and finally improves the face quality of the synthesized face. Again, this method preserves the identity features but fails to transfer the attribute parts of the target image, leading to a blurry image.\\
\vspace{-0.2cm}

\textbf{Quantitative comparison}: To compare ExtSwap quantitatively we use 3 different metrics mentioned in table \ref{tab:quant-other}. ExtSwap shows stronger identity preservation and fewer errors for pose and expression. Furthermore, ExtSwap achieved lower FID scores than the other methods.

\begin{table}[]
\centering
\caption{Quantitative comparison of ExtSwap with other method with identity similarity, pose and expression error and FID scores}
\label{tab:quant-other}
\begin{tabular}{|c|c|c|c|c|}
\hline
\textbf{Method} & \textbf{ID similarity} $\uparrow$& \textbf{Pose error}$\downarrow$ & \textbf{Exp error} $\downarrow$ &\textbf{FID} $\downarrow$  \\ \hline
HifiSwap \cite{wang2021hififace} & 0.5020 &4.688  & 3.25 &12.645  \\ \hline
Simswap \cite{chen2020simswap} & 0.5174 &4.856  &3.10  &11.98  \\ \hline
 Nitzan et al. \cite{nitzan2020face}&0.5398  &3.895  & 3.05 & 10.98 \\ \hline
 Ours& \textbf{0.5698} & \textbf{3.477} & \textbf{2.95} & \textbf{9.87}  \\ \hline
\end{tabular}
\end{table}

\vspace{-1cm}

\subsection{Comparison with CelebA-HQ dataset}
\textbf{Qualitative comparison}: We also test our model on high-quality $1024 \times 1024$ images. ExtSwap utilizes the disentangled identity and attribute features from the extended latent mapper to produce perfectly swapped faces, as shown in Fig \ref{fig:cmp_our_result_celeba}.
\vspace{-0.5cm}
\subsection{Comparison with FFHQ dataset}
\textbf{Qualitative comparison}: We perform experiments on $1024\times 1024$ high resolution FFHQ dataset. Our disentangled feature identity and attribute feature transfer led to perfect preservation of the identity and attribute features in the swapped image, as is evident in Fig \ref{fig:cmp_results_FFHQ_new}.

\section{Conclusion}
This paper presents an extended latent mapper as a novel disentangled representation. This disentangled representation acts as a key mapping technique fed into the StyleGAN to generate a quality image while requiring modest supervision. Extensive experiments demonstrate that ExtSwap can synthesize quality face images using the superiority of our disentangled identity and attribute transfer regarding hallucination, quality, and consistency of generated images.


%
%
%
%

\end{document}